\documentclass{article} %

\usepackage{colm2024_conference}

\usepackage{booktabs}
\usepackage{graphicx}
\usepackage{enumitem}
\usepackage{wrapfig}
\usepackage{algorithm}
\usepackage{algpseudocode}
\usepackage[T1]{fontenc}

\usepackage{lmodern}
\usepackage{pifont}

\usepackage{microtype}
\usepackage{amsmath}
\usepackage{colortbl}
\usepackage[utf8]{inputenc}
\definecolor{lightgray}{rgb}{0.9,0.9,0.9}
\usepackage{caption}
\usepackage{subcaption}
\usepackage{xcolor}
\usepackage{setspace}
\usepackage{url}
\usepackage{multirow}
\usepackage{colortbl}
\usepackage{tabularx}
\usepackage{blindtext}
\usepackage{pgfplots}
\pgfplotsset{compat=1.18} 
\usepackage{tikz}
\usetikzlibrary{er,positioning,bayesnet}
\usepackage{makecell}
\usepackage{tipa}
\usepackage{siunitx}
\usepackage{nicefrac}
\usepackage{tocloft}
\usepackage{listings}
\usepackage[raster,skins]{tcolorbox} %
\usepackage{xltabular}
\usepackage{adjustbox}
\usepackage{xurl}
\usepackage{marvosym}
\usepackage{algpseudocode}

\author{\textbf{Mason Kadem}
\\Computing and Software, Faculty of Engineering, McMaster University }

\title{Perception Is All You Need: A Neuroscience Framework for Low Cost Sensorless Gaze in HRI}

\begin{document}

\maketitle

\begin{abstract}
Gaze-following in child-robot interaction improves attention, recall, and learning, but requires expensive platforms (\$30,000+), sensors, algorithms, and raises privacy concerns. We propose a
framework that avoids sensors and computation entirely, instead
relying on the human visual system's assumption of convexity to produce perceptual gaze-following between a robot and its viewer. Specifically, we motivate sub-dollar cardboard robot design that directly implements the brain's own gaze computation pipeline in reverse, making the viewer's perceptual system the robot's "actuator", with no sensors, no power, and no privacy concerns. We ground this framework in three converging lines of theoretical and empirical neuroscience evidence. Namely, the distributed face processing network that computes gaze direction via the superior temporal sulcus, the high-precision convexity prior that causes the brain to perceive concave faces as convex, and the predictive processing hierarchy in which top-down face knowledge overrides bottom-up depth signals. These mechanisms explain why a concave eye socket with a painted pupil produces the perception of mutual gaze from any viewing angle. We derive design constraints from perceptual science, present a sub-dollar open-template robot with parameterized interchangeable eye inserts, and identify boundary conditions (developmental, clinical, and geometric) that predict where the framework will succeed and where it will fail. If leveraged, two decades of HRI gaze findings become deliverable at population scale.

\end{abstract}

\section{Introduction}

Eye contact is fundamental to human development. Newborns preferentially orient toward faces that make direct eye contact within hours of birth~\citep{farroni2002eye}, and prefer faces with open eyes over closed eyes~\citep{batki2000innate}. The ability to detect and follow another's gaze precedes and predicts language acquisition, attention regulation, and theory of mind~\citep{brooks2015connecting, mundy2007attention}. The capacity to share a focus of attention with another person is among the strongest early predictors of academic outcomes~\citep{loe2007academic}. Children who struggle with sustained attention fall behind in reading, mathematics, and social development, and these consequences are not equally distributed. Children in under-resourced schools and low-income families face both higher prevalence of attention difficulties and lower access to intervention.

Robot gaze can help. Gaze-following in child-robot interaction improves attention~\citep{willemse2018robot, admoni2017social}, recall~\citep{szafir2012attention}, and learning~\citep{belpaeme2018social}. But every robot that maintains eye contact with a child does so through some combination of face detection, position computation, and motorized head or eye actuation. Even minimal implementations add tens of thousands of dollars, dependency on electricity, ongoing maintenance, and a device physically capable of recording images of children. The field's most-used platforms cost over \$30,000~\citep{belpaeme2018social}, and the findings obtained on them remain confined to well-resourced labs. 

\begin{figure}[hbtp]
\centering
\includegraphics[width=0.5\textwidth]{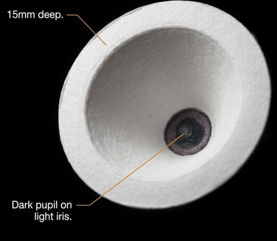}
\caption{The proposed eye}
\label{fig:eye}
\end{figure}

The question is whether sensor-based gaze is the only path, or whether the perceptual effect itself can be achieved by other means. In this work, we propose an approach that avoids sensors and computation entirely, instead relying on the human visual system to produce perceptual gaze-following between a robot and its viewer. The mechanism is the hollow-face illusion, first documented in the 1970s by Gregory~\citep{gregory1973hollow}, occurs because the brain maintains a strong prior that faces are convex. A concave face appears to rotate and track the viewer as they move. The geometric reversal inherent in this misperception means that as the viewer moves, the illusory convex face appears to counter-rotate, its eyes seem to follow. No camera detects the viewer. No motor rotates the eyes. The viewer's perceptual system is the actuator. We apply this principle to robot design with two concave eye sockets. It costs under one dollar, prints on sheets of cardstock, requires no electricity, and is physically incapable of collecting data. 

We make four contributions. First, we ground geometry-based gaze in a neuroscience framework that explains why it works, when it will fail, and what parameters matter. Second, we introduce geometry-based gaze as a new modality for HRI, distinct from sensor-based gaze in cost, privacy, and deployment. Third, we present an open sub-dollar platform with parameterized eye geometry (Fig~\ref{fig:eye}). Fourth, we derive testable predictions and boundary conditions from the framework, along with a candidate study design for empirical validation.


\section{Perceptual Foundations}
\label{sec:foundations}

The framework depends on three properties of the human visual system: a distributed network that computes gaze direction from face-like stimuli, a high-precision prior that faces are convex, and a predictive processing hierarchy in which that prior overrides conflicting sensory evidence. This section reviews each.

\subsection{The face processing network}

Face perception depends on a distributed cortical network comprising a core system of three bilateral occipitotemporal regions and an extended system that recruits additional structures for social and emotional processing \citep{haxby2000}. The core system includes the occipital face area (OFA), which performs early structural encoding of face parts; the fusiform face area (FFA), which processes invariant features underlying identity recognition \citep{kanwisher2006}; and the posterior superior temporal sulcus (STS), which processes changeable aspects of faces (e.g., gaze direction, expression, and lip movement) that facilitate social communication. The extended system includes the amygdala and insula (emotional significance), the intraparietal sulcus (spatial attention shifts triggered by observed gaze), and anterior temporal regions (biographical knowledge about recognized individuals).

For geometry-based gaze, the STS pathway is the critical channel. The STS operates in parallel with the FFA identity pathway, meaning gaze direction processing does not require recognizing \emph{who} is looking, only that something face-like \emph{is} looking. This distinction matters. The cardboard robot need not be recognized as a specific individual. It need only be parsed as sufficiently face-like to engage the STS gaze computation.

\subsection{How the visual system computes gaze direction}

Gaze direction is encoded by functionally discrete cell populations in the STS, each tuned to a specific direction. Single-cell recordings in macaque STS identified these populations organized in patches of approximately 3--5~mm, selectively responding to direct, leftward, or rightward gaze and integrating information from gaze direction, head orientation, and body posture \citep{perrett1992}. Human fMRI adaptation experiments confirmed dissociable neural representations for left versus right gaze in the anterior STS and right inferior parietal lobule \citep{calder2007}.

The visual system relies on luminance contrast between sclera and iris as the primary gaze cue. Humans possess a unique eye morphology. A large, uniformly white sclera contrasting with a dark iris, proposed to have evolved specifically for gaze signaling \citep{kobayashi1997}. Reversing this contrast polarity dramatically impairs gaze judgments \citep{ricciardelli2000}, and asymmetric darkening of one side of the sclera shifts perceived gaze direction by 8--15 degrees \citep{ando2002}. The robot's painted pupil exploits this same dark-on-light heuristic.

\begin{figure}[hbtp]
\centering
\includegraphics[width=\textwidth]{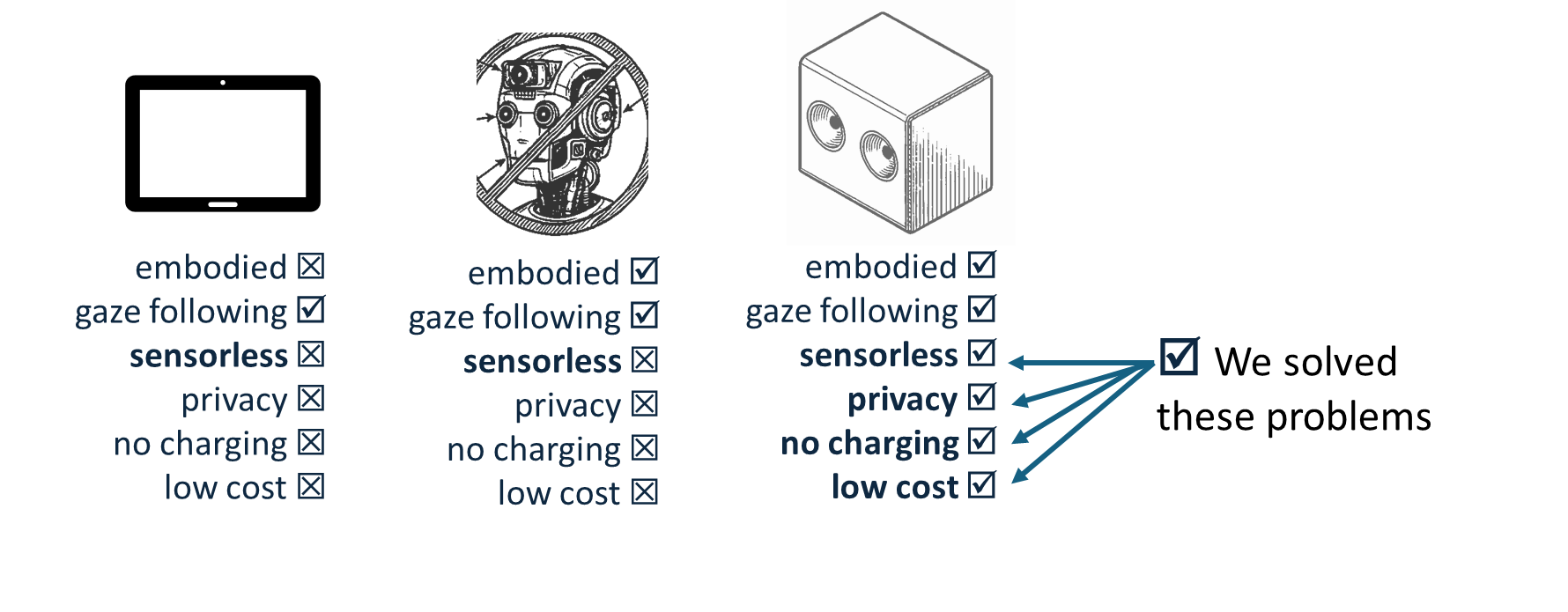}
\caption{Solved these problems}
\label{fig:compare}
\end{figure}

A subcortical route through the superior colliculus, pulvinar, and amygdala provides rapid, coarse detection of whether someone is looking at you and reflexive social attention~\citep{senju2009}. If the hollow-face percept produces the appearance of direct gaze, a fast-track system would engage, potentially triggering the full cascade of social-cognitive effects associated with mutual gaze (i.e., enhanced face encoding, amygdala activation, and increased social attention).

\subsection{The convexity prior}

The hollow-face illusion occurs when the concave interior of a face mask is perceived as a normal convex face, even against unambiguous binocular disparity cues. The phenomenon was first characterized as a case of perception operating as hypothesis-testing. When top-down knowledge about faces is pitted against bottom-up sensory signals, knowledge wins \citep{gregory1973hollow}. The face convexity prior is so strong it overrides monocular depth cues (shading, texture gradients) and even binocular stereoscopic information signaling concavity.

Three lines of evidence clarify what makes this prior face-specific rather than merely a generic convexity bias. First, the depth-inversion illusion is significantly weakened when hollow faces are presented upside-down but not when non-face reverse-perspective scenes are inverted, indicating that the spatial configuration of faces contributes substantially \citep{papathomas2004}. Second, systematic experiments have shown that familiar shading patterns and upright orientation independently strengthen the illusion, confirming that face-likeness, not just shape ambiguity, drives the percept \citep{hill2007hollow}. Third, neuroimaging has demonstrated that viewing hollow faces strengthens top-down connectivity from parietal cortex to lateral occipital cortex in healthy observers, while bottom-up connections remain unchanged \citep{dima2009hollow}. This top-down modulation is the neural signature of the prior in action.

\subsection{Developmental boundaries}

Infants under 8 months do not perceive the illusion, implying that the convexity prior develops during childhood as face experience accumulates~\citep{tsuruhara2011hollow}. FFA specialization is protracted. The right FFA in children aged 7--11 is approximately one-third the volume of the adult FFA, with expansion correlating with improved face recognition memory \citep{golarai2007}. Neonates have a preference for eye gaze~\citep{farroni2002eye}. Gaze-following emerges around 10 months \citep{brooks2005}. Joint attention is established by 18 months~\citep{mundy2007attention}. These milestones suggest the perceptual machinery may be in place by the preschool years.

\subsection{Predictive processing}

The predictive processing framework provides the most coherent theoretical account of the mechanism. Under this framework, perception is Bayesian inference. The brain maintains hierarchical generative models that produce top-down predictions, compared at every cortical level with bottom-up sensory signals \citep{friston2005}. Forward connections carry prediction errors; backward connections carry predictions. The brain minimizes prediction error by either updating its model or when the prior is sufficiently precise overriding the sensory evidence.

The face convexity prior has extremely high precision, acquired through a lifetime of encountering only convex faces. When a face-shaped stimulus is encountered, the hypothesis ``this is a convex face'' generates fewer high-level prediction errors than the veridical concave interpretation, so the illusory percept dominates \citep{hohwy2013}. The brain continuously estimates the context-dependent reliability of its own prediction errors (a process called precision-weighting \citep{clark2013}), which is reflected in physiological markers such as pupil dilation during cognitively demanding tasks \citep{kadem_pupil_2020}. Notably, this signal is robust enough to be detected even with consumer-grade webcams, suggesting that the underlying cognitive mechanisms are sufficiently strong to overcome noisy measurement channels \citep{kadem_pearls_2017}. This view of perception as hierarchical inference has a computational parallel. Recent work in mechanistic interpretability of transformer models uses attention head intervention to reveal how specific layers encode and weight features. A methodology conceptually analogous to how illusions like the hollow-face mask reveal the weighting of the face prior in the visual hierarchy~\citep{kadem_interpreting_2026}. The hollow-face illusion occurs precisely because the face prior's precision massively outweighs the precision assigned to conflicting depth signals.

The geometric consequence is that the illusory convex face appears to counter-rotate as the viewer moves. The robot's concave eye sockets exploit this. The viewer's own perceptual system acts as the gaze actuator. The framework also predicts its own boundary conditions. The effect should weaken for any population or condition where the face prior has lower precision. Infants whose prior is still developing, individuals with schizophrenia whose top-down processing is disrupted, potentially autistic individuals whose priors may be attenuated, and situations where strong conflicting cues (close binocular viewing, motion parallax at short distances) increase the precision of bottom-up error signals.

\begin{figure*}[hbtp]
\centering
\adjustbox{%
    trim=0.36\width{} 0 0 0,   
    clip,                       
    width=0.6\textwidth,        
    decodearray={1 0 1 0 1 0}   
}{%
    \includegraphics[width=\textwidth]{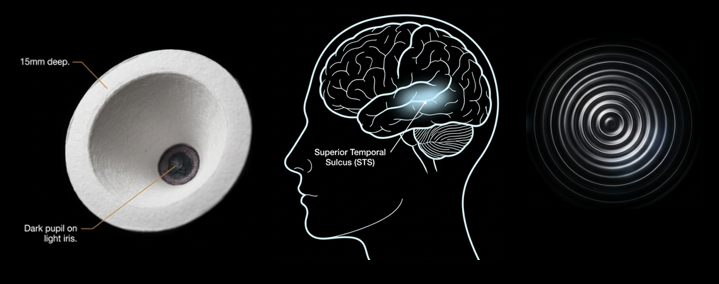}%
}
\caption{The perceptual pipeline}
\label{fig:pipeline}
\end{figure*}

The perceptual processing pipeline from retinal input to the illusory gaze-following percept. The key insight is that the robot contributes only the physical stimulus (a concave eye socket with a painted pupil). Every subsequent processing step from face detection, gaze computation, depth inversion, to the resulting social-cognitive cascade is performed by the viewer's own neural architecture.

\subsection{Clinical Disruptions}

The illusion is weakened in schizophrenia~\citep{dima2009hollow}. Patients miscategorize hollow faces as convex only approximately 6\% of the time, compared to 99\% in healthy controls. Neuroimaging shows that healthy controls show strengthened top-down connectivity from intraparietal sulcus to lateral occipital cortex during hollow-face viewing, while patients show the opposite weakened top-down and strengthened bottom-up connectivity. For autism, the influential hypo-priors hypothesis proposes that attenuated Bayesian priors lead to more veridical, less context-dependent perception \citep{pellicano2012}, predicting reduced susceptibility to the hollow-face illusion. While not yet tested in this illusion, similar disruptions in top-down predictive processing are observed in Alzheimer's disease dementia, suggesting a potential application of this low-cost paradigm for cognitive screening~\citep{kadem_interpretable_2023}. The convergence across development (absent in infants), schizophrenia (reduced with weakened top-down processing), and autism (predicted to be reduced) strongly supports the interpretation that the illusion depends on intact, mature, face-specific top-down processing and a potential opportunity for our robot design to do early screening.

\section{Related Work}
Two decades of research have converged on a consistent finding. Gaze drives engagement in child-robot interaction. Mutual gaze and gaze-following improve interaction quality~\citep{admoni2017social}, robot faces that follow gaze produce longer sustained attention than static eyes~\citep{willemse2018robot}, and adaptive gaze improves recall~\citep{szafir2012attention}. \citet{belpaeme2018social} concluded that embodied physical robots produce stronger engagement and learning gains than screen-based agents, with gaze as a key factor but noted that the high cost of platforms limits real-world applicability. The translation gap between these findings and deployment is the central motivation for the present work.

Gregory~\citep{gregory1973hollow} demonstrated that concave face masks are reliably perceived as convex by adult observers. The brain's face-convexity prior overrides depth cues, causing painted features to appear to track the viewer. \citet{hill2007hollow} found that the illusion's strength scales with concavity depth and is stronger for faces than non-face objects, suggesting a face-specific contribution beyond a general convexity bias.

The closest prior work used a mechanically switchable eye to toggle between concave and convex surfaces in a robot face~\citep{kinoshita2017transgazer}. Adult participants rated the mutual gaze eye robot as more attentive and warm. However, their system required electromechanical actuation, tested only adults, and used a single fixed geometry. Our work is the first framework to eliminate all sensing and actuation from gaze entirely, the first to provide design constraints from perceptual science, and the first to frame the approach within a neuroscience-grounded theoretical and empirical perceptual framework.

\section{Robot Design}

\begin{figure*}[hbtp]
\centering
\includegraphics[width=0.6\textwidth]{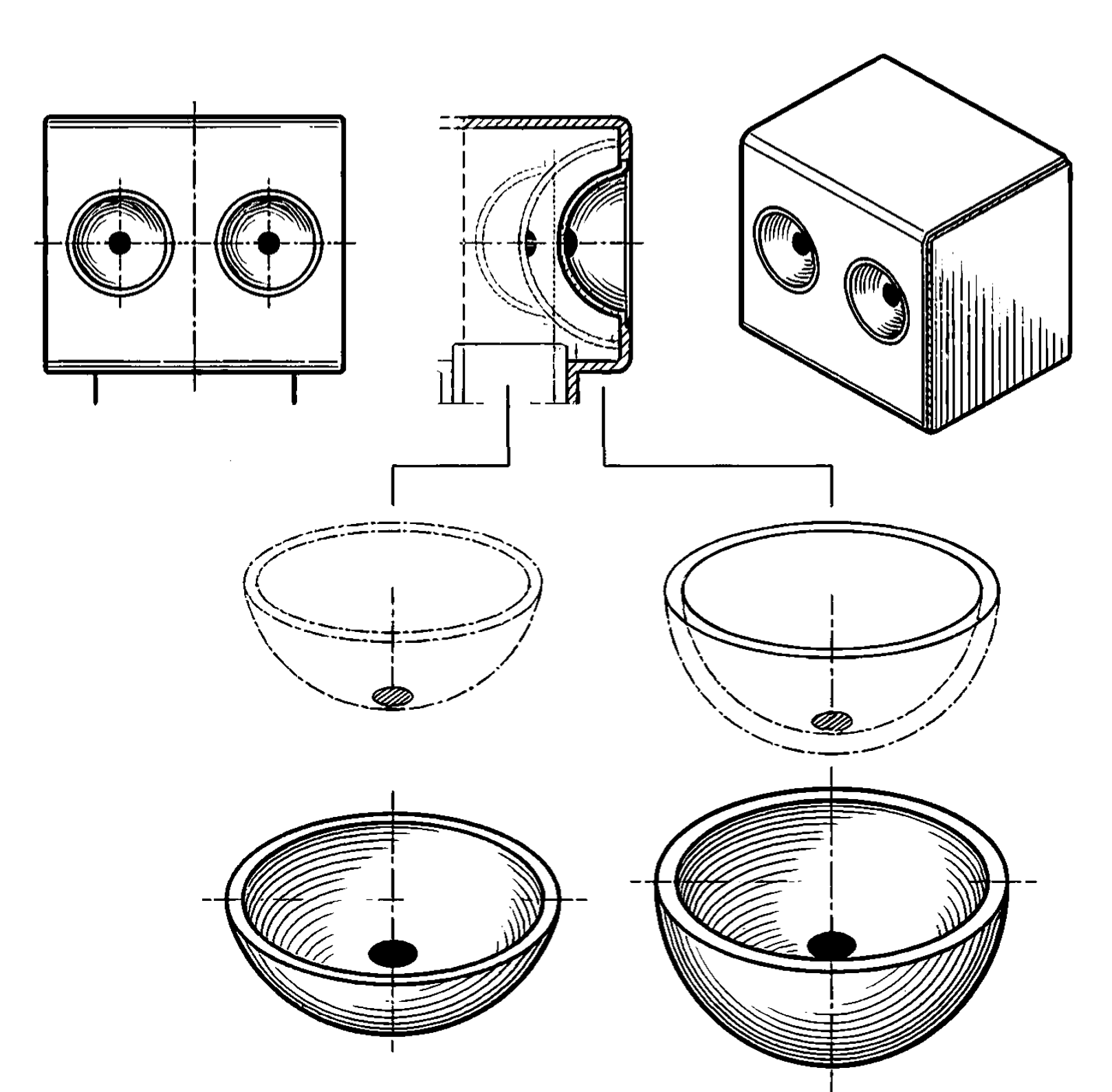}
\caption{Physics-based geometric robot eye gaze design. Concave eye inserts with painted pupils produce perceived gaze-following through the hollow-face illusion.}
\label{fig:design}
\end{figure*}

The robot is a modular body constructed from printed cardstock (Figure~\ref{fig:design}). It prints on A4 sheets and assembles in under 10 minutes with scissors and glue. Total cost is under \$1 CAD. The body template is character-agnostic. The face plate, snout, and eye apertures are fixed geometry, while the surrounding body can be reskinned as any character. The face may feature a protruding snout extending forward from the eye plane, providing a convex reference frame that reinforces the brain's expectation of facial convexity and strengthens the illusion. Two circular apertures receive interchangeable concave eye inserts from behind.

\subsubsection*{Eye geometry}
Each eye insert is formed from a flat circular disc ($\approx$ 80\,mm diameter) of printed cardstock bearing an iris and pupil. A wedge-shaped sector is removed and the remaining disc is curled into a cone. The pupil sits at the deepest point of the resulting concavity. Cone depth $d$ is determined by disc radius $R$ and wedge angle $\theta$:
\begin{equation}
    d = R\sqrt{1 - \left(\frac{360-\theta}{360}\right)^{\!2}}
\end{equation}
We construct two sets: a shallow set ($d \approx 15$\,mm) and a deep set ( $d \approx 35$\,mm). All other parameters are held constant: aperture diameter ($\approx$30\,mm), pupil diameter ($\approx$12\,mm), iris diameter ($\approx$24\,mm), pupil offset ($\approx$5\,mm toward the center), and surrounding face geometry. Inserts attach with removable tape, allowing swaps in under one minute.

\subsection{Design constraints} The neuroscience framework herein motivates several design choices (Table~\ref{tab:constraints}):

\begin{table}[hbtp]
\centering
\caption{Design constraints derived from perceptual neuroscience.}
\label{tab:constraints}
\begin{tabularx}{\textwidth}{@{} l X @{}}
\toprule
\textbf{Constraint} & \textbf{Rationale} \\
\midrule
Dark pupil on light iris & The visual system uses a dark-on-light heuristic for gaze computation \citep{ricciardelli2000}. Reversing contrast impairs or eliminates the gaze percept. \\
Face-like configuration & The convexity prior is face-specific \citep{papathomas2004,hill2007hollow}. Two eyes in a face-like arrangement with a protruding snout maximize prior engagement. An isolated concave disc without facial context produces a weaker illusion. \\
Upright orientation & The illusion is disrupted by inversion \citep{papathomas2004}. The robot must be presented upright. \\
Viewing distance $>0.5$\,m & At close range, binocular disparity provides high-precision depth information that can override the prior. The illusion is strongest where binocular cues are weak. \\
Familiar lighting & The illusion is strengthened by shading from above and weakened by unusual lighting that disrupts expected illumination gradients \citep{hill2007hollow}. \\
\bottomrule
\end{tabularx}
\end{table}

\subsubsection*{Privacy as a physical property}
A camera-based social robot can be made compliant with privacy regulations. It cannot be made incapable of surveillance, software can be updated, firmware compromised, and data policies changed. The only robot physically incapable of collecting visual data is one with no visual sensor. Our robot has no camera, no microphone, no sensor of any kind~\ref{fig:compare}. Its gaze is produced by perceptual geometry encoded by the brain's own gaze machinery, not by processing images. A device with no data collection capability does not trigger COPPA, GDPR, or equivalent requirements. Parents who would not place a camera-equipped robot in their child's room may accept a cardboard character with painted eyes. Schools and clinics facing strict rules about recording children can deploy it without legal review.

\section{Applications}

To illustrate how the sensorless design can be deployed, we describe a caregiver-in-the-loop attention support protocol. A parent or teacher connects an output only, microphone-free Bluetooth speaker inside the robot and plays pre-recorded stories. The caregiver pauses speech when the child's attention drifts and resumes when it returns, leveraging human judgment rather than replacing it with sensors. This approach aligns with emerging trends in human-centered ambient sensing for care environments, which prioritize unobtrusive monitoring over invasive data collection \citep{kadem_human-centered_2026}. This division of labor where the robot provides the persistent perceptual cue and the human provides the adaptive intelligence mirrors frameworks for effective human-clinical AI collaboration, in which the system supports rather than replaces human expertise \citep{kadem_human-clinical_2025}. The persistent perceived gaze gives the pause social weight. The robot appears to watch and wait.

The approach generalizes beyond attention support. A robot that appears to make eye contact with every viewer simultaneously could serve as a reading companion, classroom engagement aid, speech therapy prompt, waiting-room anxiety reduction tool, or early intervention resource in remote communities. The geometry-based gaze modality opens a design space that sensor-based systems cannot enter at this cost and privacy profile. 

Crucially, the underlying principle of leveraging the brain's perceptual priors to create a social signal without sensors extends beyond gaze. Any robust perceptual illusion that depends on a strong prior could, in principle, be implemented physically. For instance, the McGurk effect demonstrates that conflicting visual cues can override auditory speech perception. A physical object (e.g., lip movements) designed to exploit such cross-modal priors might improve robot-human communication without any electronics. The present framework for gaze is a specific instance of a broader design philosophy of treating the human visual system as part of the robot's control loop.
\section{Framework Validation and Future Work}

The long term objective is to leverage physics-based eye geometry to optimize functional social gaze in child-robot interaction and improve accessible translation of HRI research findings. The framework generates testable predictions. We outline the key research questions and a candidate study design for future empirical validation

The neuroscience framework generates testable predictions organized around three research questions.

\subsubsection*{To what extent do children ages 3--11 perceive the gaze-following effect, and how does perception vary with age?}
The framework predicts a developmental gradient. The convexity prior is absent in infants under 8 months \citep{tsuruhara2011hollow} and robust in adults, with FFA maturation continuing through adolescence \citep{golarai2007,scherf2007}. Children at the younger end of the range (age 3) may show weaker or absent illusion perception, while older children (age 11) should approach adult-like susceptibility. Gaze detection itself is in place by age 3 \citep{farroni2002eye}, so any age-related differences would reflect the convexity prior specifically, not gaze processing capacity.

\subsubsection*{How does concavity depth (15~mm vs.\ 35~mm) affect perceived gaze strength in children?}
Deeper concavities produce stronger illusory convexity in adults \citep{hill2007hollow}. Whether this scales linearly and whether children show the same depth sensitivity is unknown. The predictive processing framework predicts that deeper concavities generate a larger discrepancy between veridical depth and the prior's prediction, but also provide stronger bottom-up depth cues that could partially resist the prior. The optimal depth may differ between children and adults.

\subsubsection*{How concordant are identical twins' responses to geometry-based gaze?}
If the convexity prior's developmental trajectory has a heritable component, identical twins should show more concordant responses than would be expected by chance. This provides a within-pair control for genetics and shared environment that is rare in child HRI research.

\section{Conclusion}

Three streams of neuroscience evidence converge to ground the geometry-based gaze framework. The brain processes faces through a distributed network where the STS encodes gaze direction. A powerful convexity prior, shaped by lifetime experience and mediated by top-down feedback from parietal to occipital cortex causes the brain to perceive concave faces as convex. This same geometric reversal produces the illusory gaze-following effect the framework exploits.

The prior strengthens developmentally. Absent in young infants, maturing alongside FFA specialization through adolescence, and weakens when top-down processing is disrupted in schizophrenia and potentially autism. Predictive processing theory provides the unifying computational and theoretical account that the high-precision face prior overwhelms low-precision depth signals. A principle the sub-dollar cardboard robot exploits without any sensors at all. Most HRI systems treat gaze as an engineering problem. Detect the user, compute the angle, rotate the motor. The framework proposed herein treats gaze as a perceptual problem. Present a stimulus that the brain's own machinery will interpret as mutual gaze.

The most novel insight is that this framework does not merely use an illusion as a gimmick. It directly implements the brain's own gaze computation pipeline in reverse, making the viewer's perceptual system the robot's "actuator." This inversion from engineering the robot to directly implementing the brain's own gaze computation pipeline in reverse may generalize beyond gaze. Any social signal that depends on a strong perceptual prior (facial expression, biological motion, animacy) could potentially be produced by geometric or material properties rather than sensors and actuators.

If we leverage the geometry-based gaze herein, something important follows. Every HRI finding that gaze matters becomes deliverable to any family with a printer and scissors. The field has spent two decades establishing that eye contact between robots and children produces real benefits. Perceptual geometry can deliver those benefits at population scale.

\bibliographystyle{plainnat}
\bibliography{custom}

\end{document}